\documentclass[runningheads]{llncs}

\usepackage{graphicx}
\usepackage[colorlinks = true,
            breaklinks = true,
            linkcolor = black,
            urlcolor  = blue,
            citecolor = blue,
            anchorcolor = blue]{hyperref}
\usepackage{breakcites}
\usepackage{url}

\usepackage{breakurl}
\begin{document}
\title{\Large Automatic Detection of Sexist Statements Commonly Used at the Workplace\thanks{Supported by Stanford University and ISEP.}}
\titlerunning{Automatic Detection of Sexist Statements}
\author{Dylan Grosz\inst{1,2}
\and Patricia Conde-Cespedes\inst{2}}
\institute{Stanford University, Stanford CA 94305, USA\\
\email{dgrosz@stanford.edu} \and
ISEP (Institut supérieur d'électronique de Paris), Paris, FR
\email{patricia.conde-cespedes@isep.fr}}
\maketitle              % typeset the header of the contribution
\begin{abstract}
\small\baselineskip=9pt Detecting hate speech in the workplace is a unique classification task, as the underlying social context implies a subtler version of conventional hate speech. Applications regarding a state-of-the-art workplace sexism detection model include aids for Human Resources departments, AI chatbots and sentiment analysis. Most existing hate speech detection methods, although robust and accurate, focus on hate speech found on social media, specifically Twitter. The context of social media is much more anonymous than the workplace, therefore it tends to lend itself to more aggressive and “hostile” versions of sexism. %As Glick \& Fiske outline in their seminal Ambivalent Sexism Theory, there exists a distinction between “hostile” and “benevolent” sexism: the former is aggressive and degrading while the latter is subtler and superficially more positive in nature. 
Therefore, datasets with large amounts of “hostile” sexism have a slightly easier detection task since “hostile” sexist statements can hinge on a couple words that, regardless of context, tip the model off that a statement is sexist. In this paper we present a dataset of sexist statements that are more likely to be said in the workplace as well as a deep learning model that can achieve state-of-the art results. Previous research has created state-of-the-art models to distinguish “hostile” and “benevolent” sexism based simply on aggregated Twitter data. Our deep learning methods, initialized with GloVe or random word embeddings, use LSTMs with attention mechanisms to outperform those models on a more diverse, filtered dataset that is more targeted towards workplace sexism, leading to an F1 score of 0.88.

\end{abstract}

\section{Introduction}
Not long ago, women in the U.S. were not entitled to vote, yet in 2016 the first woman in history was nominated to compete against a male opponent to become President of the United States. A similar situation took place in France in 2017 when Marine Le Pen faced Emmanuel Macron in the runoff election. Is the gap between male and female opportunities in the workplace changing?  A recent study conducted by McKinsey and Company (2017) \cite{KRR2017}, reveals the gaps and patterns that exist today between women and men in corporate America. The results of the study reveal that many companies have not made enough positive changes, and as a result, women are still less likely to get a promotion or get hired for a senior level position. Some key findings from this study include, for instance:\\
\begin{itemize}
\item Corporate America awards promotions to males are about 30 percent higher rate than women in the early stages of their careers.
\item Women compete for promotions as often as men, yet they receive more resistance.\\
\end{itemize}

Mary Brinton \cite{BRI2017}, sociology professor at Harvard University and instructor of Inequality and Society in Contemporary Japan, points out that although men and women are now on an equal playing field in regard to higher education, inequality persists. Furthermore, some women who occupy important positions or get important achievements suffer from sexism at their workplace. One can mention, for example, the incident that took place in December 2018 during the \textit{Ballon d'Or} ceremony when host Martin Solveig asked the young Norwegian football player Ada Hegerberg, who was awarded the inaugural women's Ballon d'Or, was asked: "\textit{Do you know how to twerk?}" \cite{AAR2018}. Even more recently, a young scientist Katie Bouman, a postdoctoral fellow at Harvard, was publicly attributed to have constructed the first algorithm that could visualize a black hole \cite{CSA2019}. Unfortunately, this event triggered a lot of sexist remarks on social media questioning Bouman's role in the monumental discovery. For instance, a YouTube video titled \textit{Woman Does 6\% of the Work but Gets 100\% of the Credit} garnered well over 100K views. Deborah Vagins, member of the American Association of University Women, emphasized that women continue to suffer discrimination, especially when a woman works in a male-dominated field (the interested reader can see \cite{RES2019}\cite{ELF2019}\cite{GRI2019}\cite{MER2019}). Another relevant example is physicist Alessandro Strumia University of Pisa who was suspended from CERN (Conseil europ\'een pour la recherche nucl\'eaire) for making sexist comments during a presentation claiming that physics was becoming sexist against men. "\textit{the data doesn't lie-women don't like physics"}, \textit{"physics was invented and built by men"} were some of the expressions he used \cite{BBC2018}\cite{PAL2019}.\\

All these examples bring out that the prejudicial and discriminatory nature of sexist behavior unfortunately pervades nearly every social context, especially for women. This phenomenon leads sexism to manifest itself in social situations whose stakes can lay between the anonymity of social media (twitter, Facebook, youtube) and the relatively greater social accountability of the workplace.\\

In this paper, based on recent Natural Language Processing (NLP) and deep learning techniques, we built a classifier to automatically detect whether or not statements commonly said at work are sexist. We also manually built a dataset of sentences containing neutral and sexist content.\\

Section \ref{secRelatedWorks} presents a literature review of automatic hate speech detection methods using NLP methods. Moving on to our novel work and contributions, Section \ref{secDatasetDescription} describes the unique dataset used for our experimental results, one which we hope future research will incorporate and improve upon. Next, Section \ref{secMethod} describes the methods used for building our classifier. Then, in section \ref{secExperiment} we present the experimental results. Finally, Section \ref{secConclusionPers} presents our conclusion and perspectives of this study. \\

The code and dataset that are discussed in this paper will be available on \textit{\href{https://github.com/dylangrosz/Automatic\_Detection\_of\_Sexist\_Statements\_Commonly\_Used\_at\_the\_Workplace}{GitHub}} and \textit{\href{https://www.kaggle.com/dgrosz/sexist-workplace-statements}{Kaggle}}.
 
%She wrote .25% of the code. This is intellectually dishonest #AndrewChael wrote a majority of the code with help from Michael D Johnson. But hey, women need the lies to feel better...

%But then all the attention became a catalyst for a sexist backlash on social media and YouTube. It set off “what can only be described as a sexist scavenger hunt,” as The Verge described it, in which an apparently small group of vociferous men questioned Bouman’s role in the project. “People began going over her work to see how much she’d really contributed to the project that skyrocketed her to unasked-for fame.”

%and we wouldn’t even be talking about it if platforms like Twitter, Reddit, and YouTube didn’t allow trollish thinking to fester and spread virally.

\section{Related Works}\label{secRelatedWorks}

For many years, fields such as social psychology have deeply studied the nature and effects of sexist content. The many contexts where one can find sexism are further nuanced by the different forms sexist speech can take. In 1996 Glick and Fiske \cite{GLF96} devised a theory introducing the concept of \textit{Ambivalent Sexism}, which distinguishes between a \textit{benevolent} and \textit{hostile} sexism. Both forms involve issues of paternalism, predetermined ideas of women’s societal roles and sexuality; however, \textit{benevolent} sexism is superficially more positive and benign in nature than \textit{hostile} sexism, yet it can carry similar underlying assumptions and stereotypes. The distinction between the two types of "sexisms" was extended recently by Jha and Mamidi (2017) \cite{JHM2017}. The authors characterized hostile sexism by an explicitly negative attitude whereas they remarked benevolent sexism is more subtle, which is why their study was focused on identifying this less pronounced form of sexism.\\

Jha and Mamidi, 2017 \cite{JHM2017} have successfully proposed a method that can disambiguate between these benevolent and hostile sexisms, suggesting that there are perhaps detectable traits of each category. Through training SVM and Sequence-to-Sequence models on a database of hostile, benevolent and neutral tweets, the two models performed with an F1 score of 0.80 for detecting benevolent sexism and 0.61 for hostile sexism. These outcomes are quite decent considering that the little preprocessing left a relatively unstructured dataset from which to learn. With regards to the context presented in our research, the workplace features much more formal and subversive sexism as compared to that found on social media, so such success in detecting benevolent sexism is useful for our purpose.\\

Previous research has also found some success on creating models that can disambiguate various types of hate speech and offensive language in the social media context. A corpus of sexist and racist tweets was debuted by Waseem and Hovy (2016) \cite{WAH2016}. This dataset was further labeled as \textit{Hostile} and \textit{Benevolent} versions of sexism by Jha and Mamidi (2017) \cite{JHM2017} which Badjatiya et al. (2017) \cite{BAG2017}, Pitsilis et al. (2018) \cite{PIR2018} and  Founta et al. (2018) \cite{FOC2018} all use as a central training dataset in their research, each attempting to improve classification results with various model types. Waseem and Hovy (2016) \cite{WAH2016} experimented with simpler learning techniques such as logistic regression, yielding an F1 score of 0.78 when classifying offensive tweets. Later studies by \cite{BAG2017} experimented with wide varieties of deep learning architectures, but success seemed to coalesce around ensembles of Recurrent Neural Network (RNN), specifically Long Short-Term Memory (LSTM) classifiers. Results for these studies featured F1 scores ranging from 0.90 to 0.93 after adding in various boosting and embedding techniques.\\

For this research, several models were employed to figure out which best predicted workplace sexism given the data. While the more basic models relied on some form of logistic regression, most other tested models employed deep learning architectures. Of these deep learning models, the simplest used a unidirectional LSTM layers, while the most complex employed a bidirectional LSTM layer with a single attention layer \cite{BAC2015}, allowing the model to automatically focus on relevant parts of the sentence. Most of these models used GloVe embedding, a project meant to place words in a vector space with respect to their co-occurrences in a corpus\cite{Glove2014}. Some models experimented with Random Embedding, which just initializes word vectors to random values so as to not give the deep learning model any given "intuition" before training.\\

Among all this related research, none specifically considered the specific context of the workplace. Rather, most of them share a curated dataset of 16K tweets from Twitter in their hate speech detection and classification tasks. Given the substantial difference in datasets and contexts, our paper proposes a new dataset of sexist statements in the workplace and an improved companion deep learning method that can achieve results akin to these previous hate speech detection tasks.\\

%We achieved this by creating and analyzing a dataset of tweets that exhibit benevolent sexism. We classified tweets into ‘Hostile’, ‘Benevo- lent’ or ‘Others’ class depending on the kind of sexism they exhibit, by using Sup- port Vector Machines (SVM), sequence- to-sequence models and FastText classifier. We achieved the best F1-score using FastText classifier. Our work aims to analyze and understand the much prevalent ambivalent sexism in social media.

%Glick \& Fiske's analysis proves that although there exist positive, benevolent spins on these tenants of sexism, the tenants remain. Therefore, it remains essential to study the detection of such sexist content to be able to detect benevolent sexism in all its forms and contexts.\\

\section{Dataset Description}\label{secDatasetDescription} 
The dataset used in model training and testing features more than 1100 examples of statements of workplace sexism, roughly balanced between examples of certain sexism and ambiguous or neutral cases (labeled with a “1” and “0” respectively). Though this dataset features some sexist statements from Twitter, it differs from previous Twitter datasets in hate speech detection research. Previous Twitter datasets were collected via keywords and hashtags, which does not port well over to workplace speech since the nature of the dataset suffers greatly from:
\begin{enumerate}
    \item Over-representing rare sexist scenarios (e.g. the name Kat is regarded as sexist since she was a figure many people directed sexist comments during Season 6 of My Kitchen Rules (\#MKR)).
    \item Unnatural amplification of certain phraseology through retweeting since all collected retweets just reproduce the original tweet attached with the username of the user who retweeted.
    \item Learning Twitter-specific tokens, especially internet slang and hashtags, which should be left unlearned with respect to the workplace context.\\
\end{enumerate}
The Twitter portion of our dataset alleviates the first issue by filtering out these rare scenarios through generalizing certain tweets (e.g. many usages of "Kat" are converted to "she" or "her"). The second issue is resolved through removing duplicates of tweet bodies and preserving only the original tweet. The final issue was resolved manually by writing out or removing hashtags (the latter occurs if it happens at the end of the tweet and has no additional contextual relevance) and converting casual slang to its more formal, work-appropriate version (e.g. "u" becomes "you"). While 55\% of the dataset includes these generic tweets of "benevolent" sexism, other sources of workplace-related sexist speech are included to keep the source contexts of the workplace statements diversified in order to reduce overfitting on confounding keywords and phrase constructions:

\begin{itemize}
    \item 55\% - A manually filtered subset of a Twitter hate speech dataset created by \cite{WAH2016}
    \item 25\% - A manually filtered subset of work-related quotes \cite{GO2018}
    \item 20\% - Miscellaneous press quotes and faculty/student submissions\cite{Press1} \cite{Press2} \cite{Press3} \cite{Press4}
\end{itemize}
 \footnotesize{NOTE: Manual data selection and filtering was done by Grosz (male) and spot checked by Conde-Cespedes (female).}\\

\normalsize Examples of certain workplace sexism must be both conceivable in a workplace environment and somewhat professional in nature. The latter requirement is a bit loose since workplace sexism can include obvious and/or “hostile” sexism. Such examples include:\\
\begin{itemize}
    \item \texttt{"Women always get more upset than men."}
    \item \texttt{"The people at work are childish. it's run by women and when women dont agree to something, oh man."}
    \item \texttt{"I'm going to miss her resting bitch face."}
    \item \texttt{"Seeing as you two think this is a modelling competition, I give you two a score of negative ten for your looks."}\\
\end{itemize}
  
Examples of ambiguous or neutral cases include:\\
\begin{itemize}
    \item \texttt{"No mountain is high enough for a girl to climb."}
    \item \texttt{"The Belgian bar near the end of the road was a great spot to go after work"}
    \item \texttt{"It seems the world is not ready for one of the most powerful and influential countries to have a woman leader. So sad."}
    \item \texttt{"Can you explain why what she described there is wrong?"}\\
\end{itemize}

Some ethical concerns can arise in implicitly defining sexism via these datapoints. Since sexism is mostly directed towards women in the collected data, subsequent modelling will reflect that imbalance through having a more nuanced understanding and a higher confidence in labelling new examples of women-directed sexism than man-directed sexism. As a neutral counterweight to the bias, a good proportion of positive and negative examples are generic enough to detect a woman be sexist towards a man. For example, the model detects a "he” and a "she" in the statement "He thinks she should consult her gender before working here." An ideal model would give less weight to the order of the subjects, but should be able to deduce that if the predicate of the statement is somewhat negative and is paired with a he vs. she set-up, the model will lean more towards predicting the phrase as being sexist regardless of to which gender the statement was levied.\\

Of the more than 1100 total statements, 55\% are labeled as sexist ("1") while 45\% are labeled as ambiguous/neutral ("0"). The dataset is publicly available on \textit{\href{https://www.kaggle.com/dgrosz/sexist-workplace-statements}{Kaggle}}. \\

\section{Description of the Classification Method}\label{secMethod}
We experimented with various classification methods to see which would yield the best results. Our models take some inspiration from previous state-of-the-art hate speech classification models. We considered four groups of model versions, denoted from V1 to V4. All of these models take in word embeddings for each word in a sentence, initialized randomly, through GloVe or through GN-GloVe (a gender debiased version of GloVe)\cite{Zhao}. After propogating through the model, outputs a binary classification pertaining to its status as sexist.\\

In each group, there are sub-versions that experiment with different sub-architectures. In total, this research considers seven model versions. In Table \ref{table:mul} we present a summary of the performance metrics for each model in terms of \textit{recall}, \textit{precision} and \textit{F1 score}.\\

\begin{itemize}

\item \textbf{Version 1} (V1) of the models (seen in Figure \ref{fig:simple}) are a class of models using non-deep learning techniques with learned embeddings, which can serve as a baseline to which deep learning models can compare. Model V1a uses GloVe word embeddings to calculate an average embedding of the statement, while Model V1b uses GloVe embeddings, but instead of calculating the average and training a logistic regression classifier like in V1a, it trains a Gradient Boosted Decision Tree classifier. These models established a baseline F1 level of around 0.83.\newpage

\begin{figure}[h]
\centering
        \includegraphics[scale=0.45]{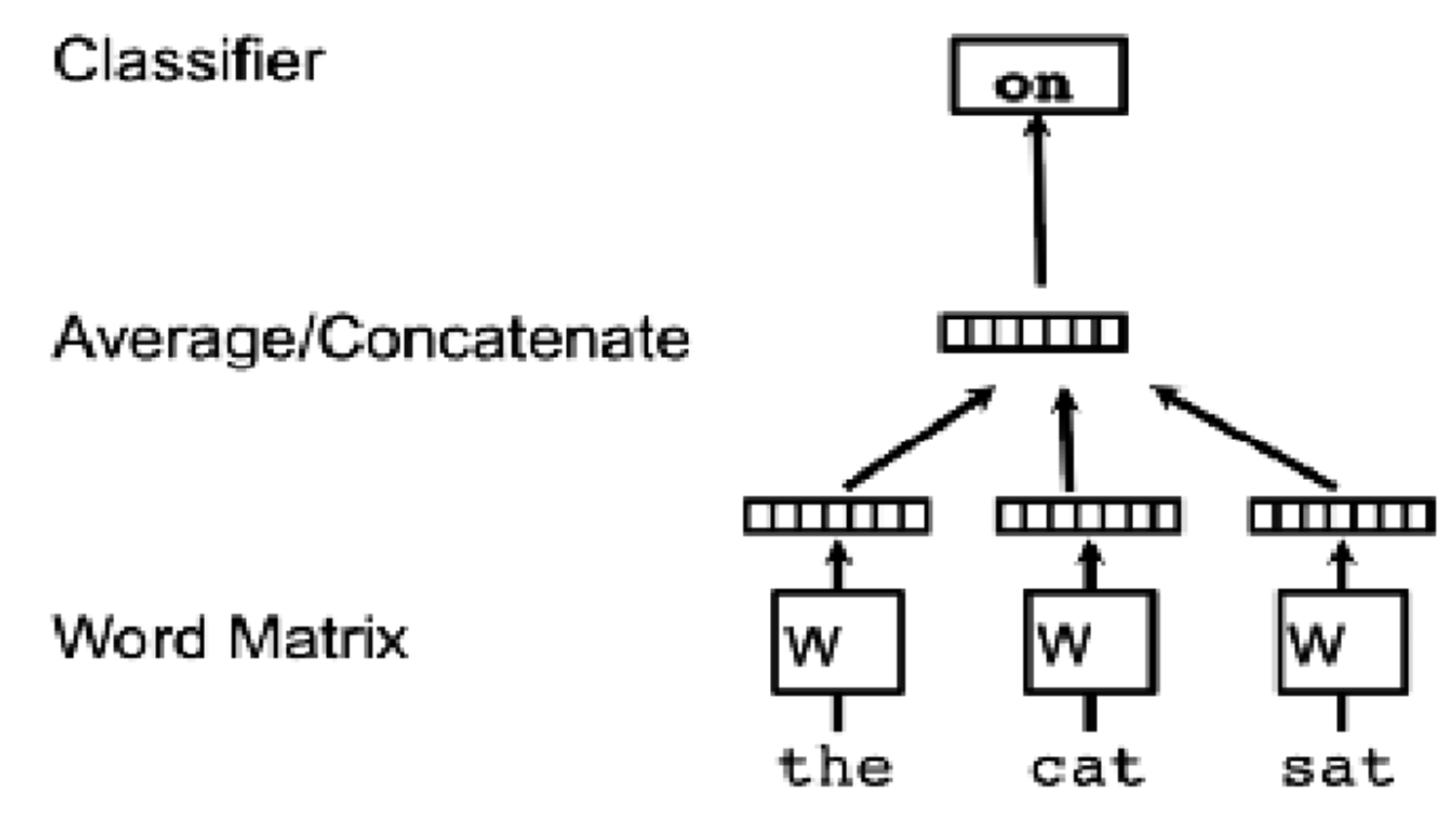}
        \caption{General architecture of models V1: a simple classifier based on word embeddings. (the image was taken from \cite{Reem2016})}
        \label{fig:simple}
\end{figure}

\item \textbf{Version 2} (V2) of the models employs a LSTM deep learning architecture (seen in Figure \ref{fig:lstm}). After an embedding layer initialized on GloVe, inputs are propagated through two unidirectional LSTM layers. In theory, this model should be able to perceive more nuanced phrases in context. For example, the V1 model would perceive a phrase such as "not pretty" individually; the LSTM construction allows the model to be able to perceive this "not pretty" as the opposite of the "pretty" in the context of its classification task. This construction had similar results to V1, also yielding a F1 of 0.83.

\begin{figure}[h!]
    \centering
    \includegraphics[scale=0.41]{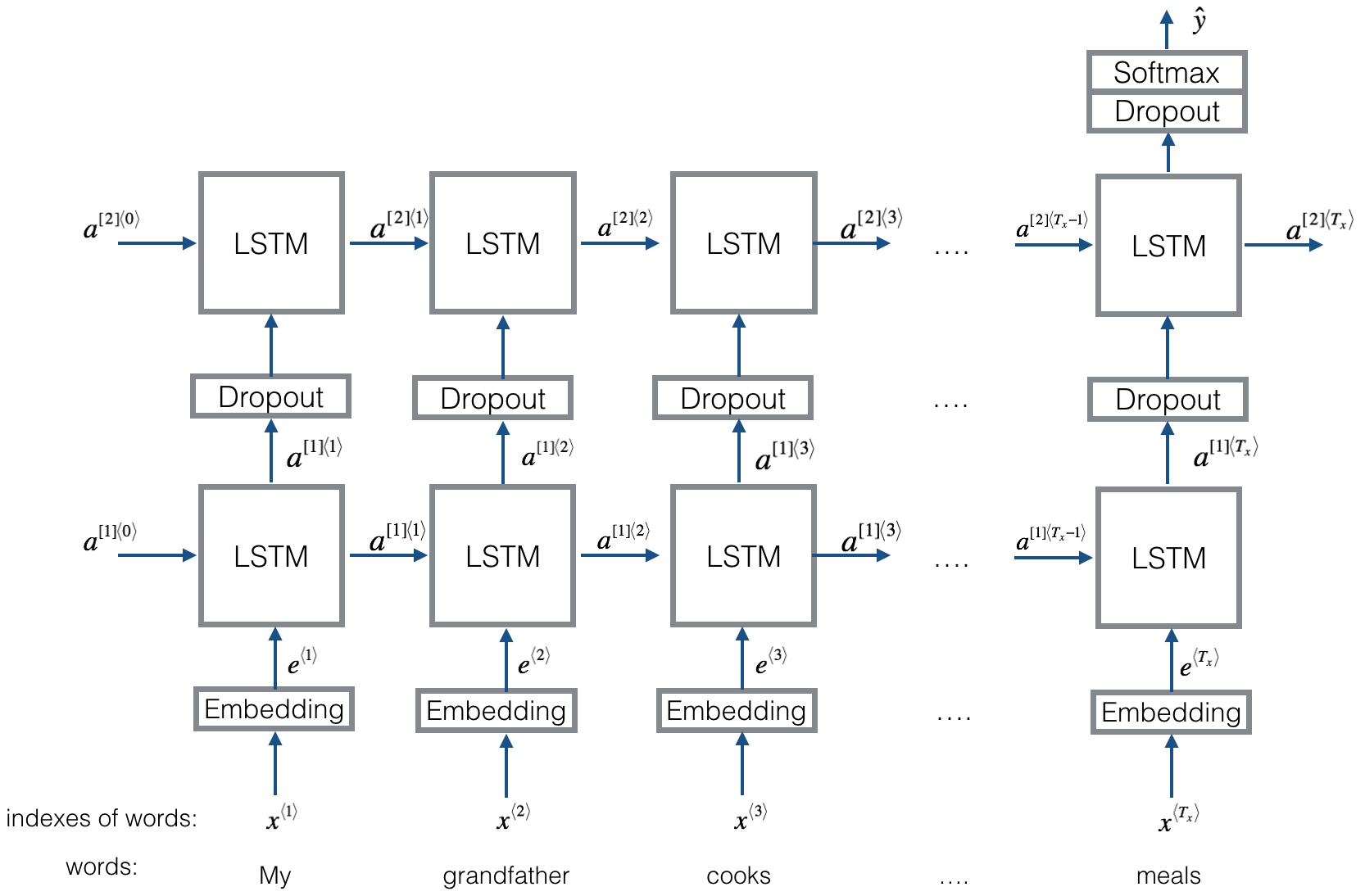}
    \caption{General architecture of models V2: A LSTM classifier based on GloVe word embeddings}
    \label{fig:lstm}
\end{figure}

\item \textbf{Version 3} (V3) of the models (seen in Figure \ref{fig:bilstm}) is very similar to V2, but it substitutes the unidirectional LSTM layers with bidirectional LSTM layers. A random embedding scheme was tested in Version 3a, GloVe for Version 3b and GN-GloVe for Version 4c. This change should allow the model to read the phrases both forwards and backwards to better learn their nuanced meanings. A phrase such as "women and men are work great together" might be more likely to be labeled as sexist by V2 due to the presence of "women and men" (which appears in many other obviously) and its ensuing influence on classification. With a separate portion of the LSTM layer devoted to "reading" the statement in the other direction, it will read "work great together" first, which will influence the classification to be non-sexist. On balance, this architecture might better perceive the nuance of certain sexist or non-sexist statements. The introduction of bidirectional layers yielded a slightly improved F1 of 0.85.\\

\begin{figure}[h]
    \centering
    \includegraphics[scale=0.65]{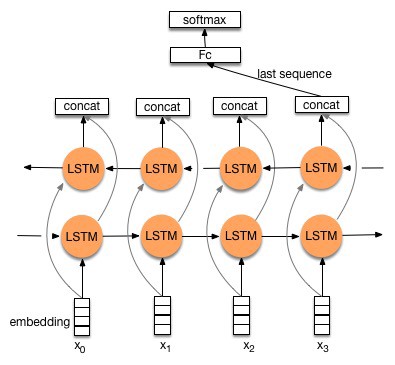}
    \caption{Architecture of the BiLSTM component of model V3: A deep learning model with a bidirectional LSTM layer that can understand sentences both forwards and backwards \cite{Fig3}}
    \label{fig:bilstm}
\end{figure}

\item \textbf{Version 4} (V4) of the models (seen in Figure \ref{fig:bilstm+attn}) employs the same architecture as V3. However, it adds a simple attention mechanism over the embedding input layer in order to focus on the significance of individual words out of context. Like in V3, random embedding was tested in Version 4a, GloVe for Version 4b and GN-GloVe for Version 4c. For example, the model tends to over-label statements including "women and men" as sexist, since it implies a comparison which usually invokes sexist stereotypes; however, there are many cases where "women and men" are followed by an undeniably neutral clause, as seen in the example statement "men and women should like this product." The attention mechanism seeks to learn that “should like this product,” usually regardless of context, means a workplace statement is not sexist. As a result of this  greater understanding of statements' nuance, this model fared best with a F1 of 0.88.

\begin{figure*}[h!]
    \centering
    \includegraphics[scale=0.35]{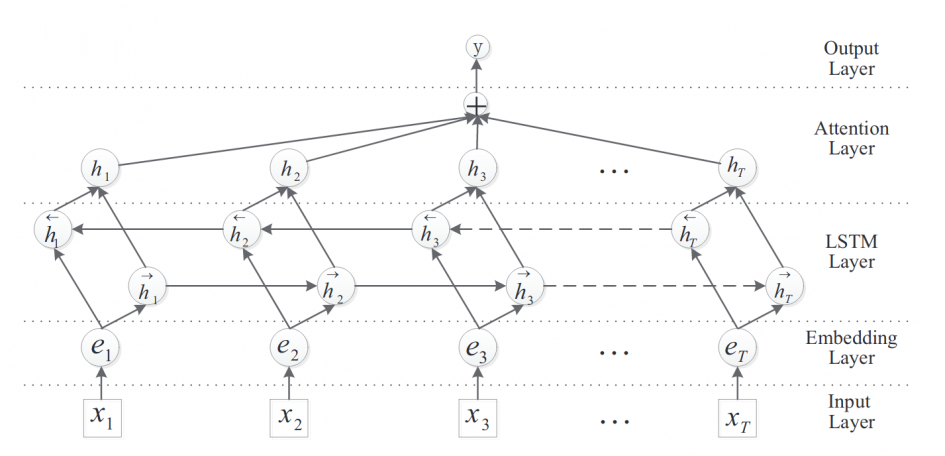}
    \caption{General architecture of the BiLSTM+Attention component of model V4: A bidirectional LSTM classifier that propagates into an attention layer \cite{Fig4}}
    \label{fig:bilstm+attn}
\end{figure*}

\end{itemize}

%\bibitem{HBMG}
%R.~E. Bank, T.~F. Dupont, and H.~Yserentant, {\em The hierarchical basis
%  multigrid method}, Numer. Math., 52 (1988), pp.~427--458.

\section{Experimental results}\label{secExperiment}
\begin{table*}[hb!]
%\begin{tabular}{|l|l|l|l|l|}
\centering
\begin{tabular}{p{0.16\linewidth}p{0.35\linewidth}p{0.12\linewidth}p{0.1\linewidth}p{0.12\linewidth}}
\hline
\textit{Model} & \textit{Description}      & \textit{Precision} & \textit{Recall} & \textit{F1 Score} \\ \hline
V1a             & GloVe+Logistic Regression & 0.81               & 0.86            & 0.83              \\
V1b             & GloVe+GBDT            & 0.82               & 0.84            & 0.83              \\ \hline

V2             & GloVe+LSTM                & 0.80                & 0.87            & 0.83              \\ \hline
V3a            & Random+BiLSTM             & 0.80                  & 0.77               & 0.79                 \\ 
V3b            & GloVe+BiLSTM              & 0.82               & 0.89            & 0.85              \\
V3c            & GN-GloVe+BiLSTM              & 0.82               & 0.89            & 0.85              \\\hline
V4a            & Random+BiLSTM+Attn        & 0.71                  & 0.81               & 0.76                 \\ 
V4b            & GloVe+BiLSTM+Attn         & \textbf{0.84}               & \textbf{0.93}            & \textbf{0.88}     \\
V4c            & GN-GloVe+BiLSTM+Attn              & 0.82               & 0.92            & 0.87              \\\hline
\end{tabular}
\caption{Model Performances} \label{table:mul}
\end{table*}

Though the simple V1 model and the unidirectional LSTM initialized on GloVe posted similar F1 scores, changing the LSTM layers to be bidirectional and adding a simple attention mechanism substantially improved the F1 score to 0.88. Though promising in previous research, initializing with random embeddings led to poor F1 scores and irreconcilable overfitting.\\

While the pretrained GloVe embedding led to the best results, a common criticism of such pretrained embeddings is the possibility that they can assume certain human biases, such as gender bias. However, training the model on a gender neutral version of GloVe (GN-GloVe) showed no significant improvement to performance\cite{Zhao}, possibly due to either a slight advantage on having mathematically embedded gender biases or the irrelevance of analogical gender biases with respect to this task. However, gender neutral word embeddings may prove promising as underlying detection tasks evolve and more research comes out regarding the debiasing of generic, complex word embeddings like GloVe, as opposed to targeted, simpler word embeddings like Google News' word2vec\cite{Boluk}. \\

Even for the best model, persistent issues include an over-aggressive labeling of sentences that include the phrase "women and men," slight overfitting despite Dropout layers and recall slightly outperforming precision (the models over-labeled statements as sexist as a whole).\\

For optimal training and testing, V2, the V3s, and the V4s featured layers with sizes between 64 and 128. There are also Dropout layers between each LSTM layer to reduce overfitting. The model was then compiled to optimize via binary cross entropy and an 'adam' optimizer.

\section{Conclusion and future works}\label{secConclusionPers}
The GloVe+BiLSTM+Attn model's F1 score of 0.88 shows that with the slightly different deep learning methods shown in this paper, a F1 score that is at the level of previous sexist detection research is attainable. This performance must also be taken into context with this task's added limitation of constraining all data to be in a workplace context; this type of data leans much more into the category of the more nuanced, subtler "benevolent" sexism.\\

With a larger dataset, the GloVe+BiLSTM+Attn will be more able to abstract from the data and learn the most generalized and accurate model possible. Although the dataset size is the most obvious culprit for being the bottleneck for further F1 improvement, there are also more possible, complex and novel deep learning architectures that can be explored and tested on the dataset, including boosting techniques, other pretrained word embedding and more sophisticated attention mechanisms. This final improvement could pose an especially ripe area for including more explainablity and understanding since attention mechanisms can allow one to peer into the key words and phrases the model focuses on when tagging statements as sexist or not.\\

The dataset presented in this paper could also be used as a basis for unsupervised learning tasks via clustering to reverse-engineer more nuanced types of sexism, as there can possibly be subclasses of both hostile and benevolent sexism that upon discovery could help sociological reframings of this problem as well as helping understand this task itself.\\

The dataset used in this research, though large enough to produce substantially robust models in workplace sexism detection, must always grow in order to capture most keywords and phrase structures found in workplace sexism. Despite the challenges posed by the current state of the dataset and model, state-of-the-art results were attained. As this dataset of sexist workplace statements grows through a crowdsourced effort, the performance of this model will improve as well.

{\small
\bibliographystyle{splncs04}
\bibliography{main.bib}

\begin{thebibliography}{10}
\providecommand{\url}[1]{\texttt{#1}}
\providecommand{\urlprefix}{URL }
\providecommand{\doi}[1]{https://doi.org/#1}

\bibitem{AAR2018}
Aarons, E.: Ada Hegerberg: First Women's Ballon d'Or Marred as Winner is Asked
  to Twerk. The Guardian (2018),
  \url{https://www.theguardian.com/football/2018/dec/03/ballon-dor-ada-hegerberg-twerk-luka-modric}

\bibitem{BAG2017}
Badjatiya, P., Gupta, S., Gupta, M., Varma, V.: Deep learning for hate speech
  detection in tweets. 26th International Conference on World Wide Web
  Companion pp. 759--760 (2017)

\bibitem{BAC2015}
Bahdanau, D., Cho, K., Bengio, Y.: Neural machine translation by jointly
  learning to align and translate. International Conference on Learning
  Representations (2015)

\bibitem{Boluk}
Bolukbasi, e.a.: Man is to computer programmer as woman is to homemaker?
  debiasing word embeddings. NIPS  (2016)

\bibitem{BRI2017}
Brinton, M.: Gender inequality and women in the workplace. Harvard Summer
  School Review  (May 2017),
  \url{https://www.summer.harvard.edu/inside-summer/gender-inequality-women-workplace}

\bibitem{Press1}
Chira, S., Milord, B.: “is there a man i can talk to?: Stories of sexism in
  the workplace.” the new york times. The New York Times  (June 2017),
  \url{www.nytimes.com/2017/06/20/business/women-react-to-sexism-in-the-workplace.html}

\bibitem{ELF2019}
Elfrink, T.: Trolls hijacked a scientist’s image to attack katie bouman. they
  picked the wrong astrophysicist. The Washington Post  (April 2019),
  \url{https://www.washingtonpost.com/nation/2019/04/12/trolls-hijacked-scientists-image-attack-katie-bouman-they-picked-wrong-astrophysicist}

\bibitem{FOC2018}
Founta, A.M., Chatzakou, D., Kourtellis, N., Blackburn, J., Vakali, A.,
  Leontiadis, I.: A unified deep learning architecture for abuse detection.
  CoRR  (2018), \url{http://arxiv.org/abs/1802.00385}

\bibitem{GLF96}
Glick, P., Fiske, S.T.: The ambivalent sexism inventory: Differentiating
  hostile and benevolent sexism. Journal of Personality and Social Psychology
  \textbf{3},  491--512 (1996)

\bibitem{GO2018}
Goel, S., M.R., Garg, S.: Proposing contextually relevant quotes for images.
  ECIR 2018: Advances in Information Retrieval  \textbf{10772} (2018)

\bibitem{GRI2019}
Griggs, M.B.: Online trolls are harassing a scientist who helped take the first
  picture of a black hole. The Verge  (April 2019),
  \url{https://www.theverge.com/2019/4/13/18308652/katie-bouman-black-hole-science-internet}

\bibitem{JHM2017}
Jha, A., Mamidi, R.: When does a compliment become sexist? Analysis and
  classification of ambivalent sexism using twitter data. Association for
  Computational Linguistics, proceedings of the second workshop on {NLP} and
  computational social science edn. (2017)

\bibitem{Reem2016}
Kabbani, R.: Checking Eligibility of Google and Microsoft Machine Learning
  Tools for use by JACK e-Learning System  (2016)

\bibitem{KRR2017}
Krivkovich, A., Robinson, K., Starikova, I., Valentino, R., Yee, L.: Women in
  the Workplace. McKinsey \& Company (2017),
  \url{https://www.mckinsey.com/featured-insights/gender-equality/women-in-the-workplace-2017}

\bibitem{Fig3}
Lee, C.: Understanding bidirectional rnn in pytorch. Towards Data Science
  (November 2017),
  \url{towardsdatascience.com/understanding-bidirectional-rnn-in-pytorch-5bd25a5dd66}

\bibitem{Press3}
McCormack, C.: 18 sexist phrases we should stop using immediately. MSN  (2018),
  \url{https://www.msn.com/en-us/Lifestyle/smart-living/18-sexist-phrases-we-should-stop-using-immediately/ss-BBLgg1E/}

\bibitem{MER2019}
Mervosh, S.: How katie bouman accidentally became the face of the black hole
  project. The New York Times  (April 2019),
  \url{https://www.nytimes.com/2019/04/11/science/katie-bouman-black-hole.html}

\bibitem{CSA2019}
@MIT\_CSAIL:  (April 2019), \url{https://twitter.com/MIT\_CSAIL/status/
  1116020858282180609}

\bibitem{BBC2018}
News, B.: Cern scientist alessandro strumia suspended after comments. BBC News
  (October 2018), \url{https://www.bbc.com/news/world-europe-45709205}

\bibitem{PAL2019}
Palus, S.: We annotated that horrible article about how women don’t like
  physics. Slate  (March 2019),
  \url{https://slate.com/technology/2019/03/women-dont-like-physics-article-annotated.html}

\bibitem{Glove2014}
Pennington, J., Socher, R., Manning, C.D.: Glove: Global vectors for word
  representation  (2014)

\bibitem{PIR2018}
Pitsilis, G., Ramampiaro, H., Langseth, H.: Detecting offensive language in
  tweets using deep learning. CoRR  (2018)

\bibitem{Press4}
Priestley, A.: Six common manifestations of everyday sexism at work.
  SmartCompany  (October 2017),
  \url{www.smartcompany.com.au/people-human-resources/six-common-manifestations-everyday-sexism-work/}

\bibitem{RES2019}
Resnick, B.: Male scientists are often cast as lone geniuses. here’s what
  happened when a woman was. Vox  (April 2019),
  \url{https://www.vox.com/science-and-health/2019/4/16/18311194/black-hole-katie-bouman-trolls}

\bibitem{WAH2016}
Waseem, Z., Hovy, D.: Hateful symbols or hateful people? predictive features
  for hate speech detection on twitter. Proceedings of the {NAACL} Student
  Research Workshop at Association for Computational Linguistics
  \textbf{2016},  88--93 (June 2016)

\bibitem{Press2}
Wolfe, L.: Sexist comments and quotes made by the media. The Balance Careers
  (March 2019),
  \url{www.thebalancecareers.com/sexist-comments-made-by-media-3515717}

\bibitem{Zhao}
Zhao, J., Zhou, Y., Li, Z., Wang, W., Chang, K.W.: Learning gender-neutral word
  embeddings. ACL pp. 4847--4853 (2018)

\bibitem{Fig4}
Zhou, P., Shi, W., Tian, J., Qi, Z., Li, B., Hao, H., Xu, B.: Attention-Based
  Bidirectional Long Short-Term Memory Networks for Relation Classification.
  ACL (2016), \url{http://www.aclweb.org/anthology/P16-2034}

\end{thebibliography}
}

\vspace{12pt}
\end{document}